\documentclass[a4paper,twoside]{article}

\usepackage{epsfig}
\usepackage{subcaption}
\usepackage{calc}
\usepackage{amssymb}
\usepackage{amstext}
\usepackage{amsmath}
\usepackage{amsthm}
\usepackage{multicol}
\usepackage{pslatex}
\usepackage{apalike}

\usepackage{times}
\usepackage{graphicx}

\usepackage{color}
\usepackage{algorithmic}
\usepackage[ruled,vlined, linesnumbered]{algorithm2e}
\usepackage[table]{xcolor}
 \usepackage{multirow}
 \usepackage{caption}
 \usepackage{bm}
 \usepackage{mathtools}
 \DeclarePairedDelimiter\floor{\lfloor}{\rfloor}
 \definecolor{lightergray}{rgb}{0.8, 0.8, 0.8}
 \usepackage{siunitx}

\usepackage[rightcaption]{sidecap}

\newcommand{\norm}[1]{\left\lVert#1\right\rVert}
\usepackage[pagebackref=true,breaklinks=true,letterpaper=true,colorlinks,bookmarks=false]{hyperref}

\usepackage{SCITEPRESS}

\begin{document}

\title{Practical Auto-Calibration for Spatial Scene-Understanding from Crowdsourced Dashcamera Videos}
\author{

\authorname{
Hemang Chawla, 
Matti Jukola,
Shabbir Marzban,
Elahe Arani,
and Bahram Zonooz}
\affiliation{Advanced Research Lab, Navinfo Europe, The Netherlands}
\email{\{hemang.chawla, matti.jukola, shabbir.marzban, elahe.arani, b.yoosefizonooz\}@navinfo.eu}
}


\abstract{
Spatial scene-understanding, including dense depth and ego-motion estimation, is an important problem in computer vision for autonomous vehicles and advanced driver assistance systems. Thus, it is beneficial to design perception modules that can utilize crowdsourced videos collected from arbitrary vehicular onboard or dashboard cameras. 
However, the intrinsic parameters corresponding to such cameras are often unknown or change over time. Typical manual calibration approaches require objects such as a chessboard or  additional scene-specific information.
On the other hand, automatic camera calibration does not have such requirements. 
Yet, the automatic calibration of dashboard cameras is challenging as forward and planar navigation results in critical motion sequences with reconstruction ambiguities. Structure reconstruction of complete visual-sequences that may contain tens of thousands of images is also computationally untenable.  Here, we propose a system for practical monocular onboard camera auto-calibration from crowdsourced videos.  
We show the effectiveness of our proposed system on the KITTI raw, Oxford RobotCar, and the crowdsourced D$^2$-City datasets in varying conditions.
Finally, we demonstrate its application for accurate monocular dense depth and ego-motion estimation on uncalibrated videos.}

\onecolumn \maketitle \normalsize \setcounter{footnote}{0} \vfill

\section{\uppercase {Introduction}}
\label{sec:introduction}
\noindent Autonomous driving systems have progressed over the years with advances in visual perception technology that enables safer driving.
These advances in computer vision have been possible with the enormous amount of visual data being captured for training neural networks applied to a variety of scene-understanding tasks such as dense depth and ego-motion estimation.
Nevertheless, acquiring and annotating vehicular onboard sensor data is a costly process.
One of the ways to design generic perception systems for spatial scene-understanding is to utilize crowdsourced data.
Unlike most available datasets which contain limited hours of visual information, exploiting large scale crowdsourced data offers a promising alternative~\cite{dabeer2017end,gordon2019depth}

Crowdsourced data is typically collected from low-cost setups such as monocular dashboard cameras.
However, the robustness of modern multi-view perception systems used in dense depth estimation~\cite{godard2018digging}, visual odometry~\cite{mur2015orb}, lane detection~\cite{ying2016robust}, object-specific distance estimation~\cite{zhu2019learning}, optical flow computation~\cite{meister2018unflow}, and so on, depends upon the accuracy of their camera intrinsics. The lack of known camera intrinsics for crowdsourced data captured from unconstrained environments prohibits the direct application of existing
approaches.
Therefore, it is pertinent to estimate these parameters, namely focal lengths, optical center, and distortion coefficients automatically and accurately. Yet, standard approaches to obtaining these parameters necessitate the use of calibration equipment such as a chessboard, or are dependent upon the presence of specific scene geometry such as planes or lines~\cite{wildenauer2013closed}.
A variety of approaches to automatically extract the camera intrinsics from a collection of images have also been proposed. Multi-view geometry based methods utilize epipolar constraints through feature extraction and matching for auto-calibration~\cite{gherardi2010practical,kukelova2015efficient}. 
However, for the typical driving scenario with constant but unknown intrinsics, forward and planar camera motion results in critical sequences~\cite{steger2012estimating,wu2014critical}.
Supervised deep learning methods instead require images with known ground truth (GT) parameters for training~\cite{lopez2019deep,zhuang2019degeneracy}. 
While Structure-from-Motion has also been used for auto-calibration, its direct application to long crowdsourced onboard videos is computationally expensive~\cite{schonberger2016structure}, and hence unscalable. This motivates the need for a practical auto-calibration method for spatial scene-understanding from unknown dashcameras.

Recently, camera auto-calibration through Structure-from-Motion (SfM) on sub-sequences of turns from KITTI raw dataset~\cite{geiger2012we} was proposed for 3D positioning of traffic signs~\cite{chawla2020monocular}.
However, the method was limited by the need for Global Positioning System (GPS) information corresponding to each image in the dataset. The GPS information may not always be available, or may be collected at a different frequency than the images, and may be noisy.
Furthermore, no analysis was performed on the kind of sub-sequences necessary for successful and accurate calibration.
Typical visibility of ego-vehicle in the onboard images also poses a problem to the direct application of SfM. Therefore, scalable accurate auto-calibration from onboard visual sequences remains a challenging problem.

In this paper, we present a practical method for extracting camera parameters including focal lengths, principal point, and radial distortion (barrel and pincushion) coefficients from a sequence of images collected using only an onboard monocular camera. Our contributions are as follows:
\begin{itemize}
    \item We analytically demonstrate that the sub-sequences of images where the vehicle is turning provide the relevant structure necessary for a successful auto-calibration.
    \item We introduce an approach to automatically determine these turns using the images from the sequence themselves.
    \item We empirically study the relation of the frames per second (fps) and number of turns in a video sequence to the calibration performance, showing that a total $\approx$ \SI{30}{s} of sub-sequences are sufficient for calibration.
    \item A semantic segmentation network is additionally used to deal with the variable shape and amount of visibility of ego-vehicle in the image sequences, improving the calibration accuracy.
    \item We validate our proposed system on the KITTI raw, the Oxford Robotcar~\cite{maddern20171}, and the D$^2$-City~\cite{che2019d} datasets against state-of-the-art.
    \item Finally, we demonstrate its application to chessboard-free dense depth and ego-motion estimation on the uncalibrated KITTI Eigen~\cite{eigen2015predicting} and Odometry~\cite{zhou2017unsupervised} splits respectively.
\end{itemize}
\section{\uppercase{Related Work}}
\label{sec:related_work}
\begin{figure*}[htbp]
\centering
\includegraphics[width=\linewidth]{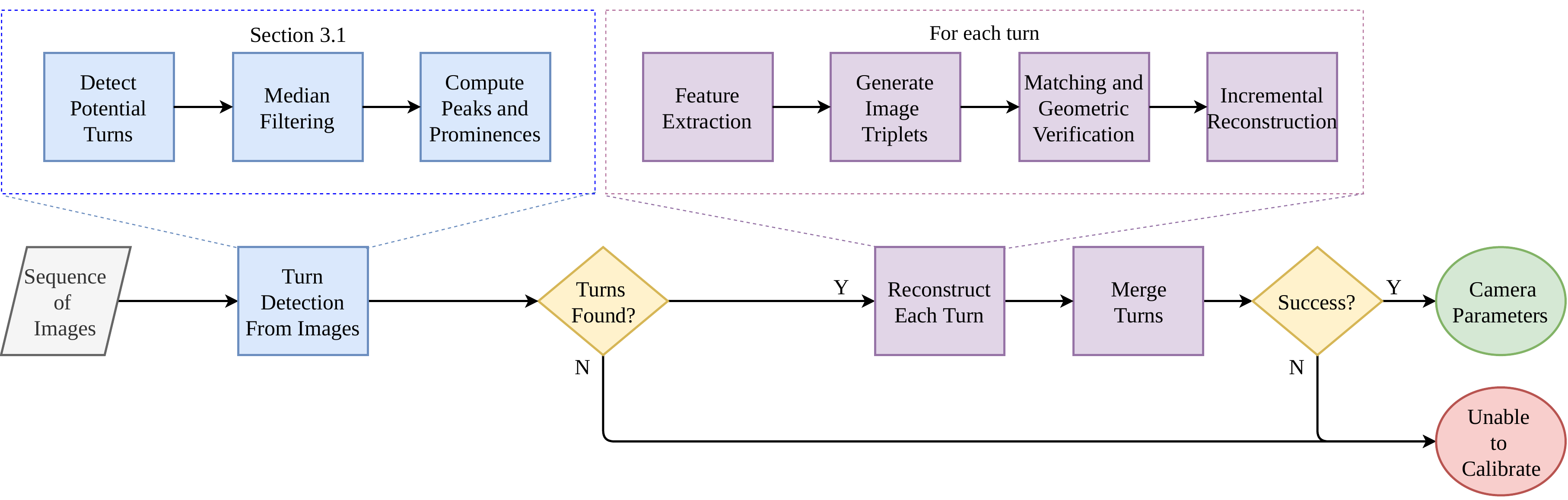}
\caption{Practical Auto Calibration from on-board visual sequences. The input to the system is represented in gray. The components of turn detection step (Section \ref{sec:turn_detection}) are shown in blue. The components of the turn reconstruction for parameter extraction (Section \ref{sec:calibration}) are shown in purple.}
\label{fig:CalibFlowchart}
\end{figure*}
\noindent Over the years, multiple ways have been devised to estimate camera parameters from a collection of images for which the camera hardware is either unknown or inaccessible.
Methods using two or more views of the scene have been proposed to estimate camera focal lengths~\cite{gherardi2010practical}.
The principal point is often fixed at the center of the image, as its calibration is an ill-posed problem~\cite{de1998self}. 
To estimate radial distortion coefficients, two-view epipolar geometry constraints are often used~\cite{kukelova2015efficient}. 
Note that forward camera motion is a degenerate scenario for distortion estimation due to ambiguity against scene depth.
Similarly, planar camera motion is also a critical calibration sequence~\cite{wu2014critical}. Nevertheless, typical automotive visual data captured from dashcameras majorly constitute forward motion on straight paths and a few turns, within a mostly planar motion. 
Supervised learning based methods for camera auto-calibration have also been introduced~\cite{lopez2019deep,zhuang2019degeneracy}. 
However, their applicability is constrained by the variety of images with different combinations of ground truth camera parameters used in training. On the other-hand self-supervised methods~\cite{gordon2019depth} do not achieve similar performance~\cite{chawla2020crowdsourced}. SfM has also been utilized to estimate camera parameters from a crowdsourced collection of images~\cite{schonberger2016structure}.
However, the reconstruction of a complete long driving sequence of tens of thousands of images is computationally expensive, motivating the utilization of a relevant subset of images~\cite{chawla2020monocular}.
Therefore, this work proposes a practical system for intrinsics auto-calibration from onboard visual sequences. Our work can be integrated with the automatic extrinsic calibration of \cite{tummala2019smartdashcam} for obtaining the complete camera calibration.
\section{\uppercase{System Design}}
\label{sec:system}

\noindent This section describes the components of the proposed practical system for camera auto-calibration using a crowdsourced sequence of $n$ images captured from an onboard monocular camera.
We represent the camera parameters using the pinhole camera model and a polynomial radial distortion model with two parameters~\cite{hartley2003multiple}. 
The intrinsic matrix is given by,
\begin{equation}
\label{eq:camera_matrix}
    K = \begin{bmatrix}
    f_x & 0 & c_x\\
    0 & f_y & c_y\\
    0 & 0 & 1
    \end{bmatrix},
\end{equation}
where $f_x$ and $f_y$ are the focal lengths, and $(c_x, c_y)$ represents the principal point. The radial distortion is modeled using a polynomial with two parameters $k_1$ and $k_2$ such that,
\begin{equation}
\label{eq:radial_distortion}
    \begin{bmatrix}
    x_d \\
    y_d
    \end{bmatrix}
     =
     (1 + k_1 r^2 + k_2 r^4)
     \begin{bmatrix}
    x_u \\
    y_u
    \end{bmatrix},
\end{equation}
where $(x_d,y_d)$ and $(x_u, y_u)$ represent the distorted and rectified pixel coordinates respectively, while $r$ is the distance of the coordinate from the distortion center (assumed to be the same as the principal point). 
An overview of the proposed framework is shown in Figure~\ref{fig:CalibFlowchart}. 

Our system is composed of two broad modules. The first module is \textit{turn detection} which outputs a list of $\varsigma$ sub-sequences corresponding to the turns in the video. The second module is \textit{incremental reconstruction based auto-calibration} using these sub-sequences. This involves building the scene graph of image pairs with geometrically verified feature matches. The scene graph is then used to reconstruct each detected turn within a bundle adjustment framework, followed by a merging of the turns. This allows to extract the single set of camera parameters from the captured turn sequences. 

Turns are necessary in extracting the camera parameters for two reasons: 
\begin{enumerate}
    \item As stated earlier, the pure translation and pure forward motion of the camera are degenerate scenarios for auto-calibration of the focal lengths and the distortion coefficients respectively~\cite{steger2012estimating,wu2014critical}.
    \item  Moreover, the error in estimating the focal lengths is bounded by the inverse of the rotation between the matched frames, as derived in Sec \ref{sec:calibration}. 

\end{enumerate}

\subsection{Turn Detection}
\label{sec:turn_detection}

Algorithm \ref{alg:turn_detection} summarizes the proposed method for turn detection. This method estimates the respective median images for the turns present in the full video. Thereafter for each median image (turn center), $k$ preceding and succeeding images are collated to form the turn sub-sequences.  

\begin{algorithm}[tbh]
\footnotesize
\DontPrintSemicolon
\SetKwInOut{Input}{input}\SetKwInOut{Output}{output}
\SetKwFunction{Associate}{associate}
\SetKwFunction{Available}{available}
\SetKwFunction{ComputePotentialTurns}{computePotentialTurns}
\SetKwFunction{MedianFilter}{medianFilter}
\SetKwFunction{FindPeaks}{findPeaks}
\SetKwFunction{Len}{len}
\SetKwFunction{Reverse}{reverse}
\SetKwFunction{Min}{min}
\SetKwFunction{Sort}{sort}
\SetKwData{I}{I}
\SetKwData{NumTurns}{$\varsigma$}
\SetKwData{G}{G}
\SetKwData{Matches}{matches}
\SetKwData{TurnCenters}{turn\_centers}
\SetKwData{PotentialTurns}{potential\_turns}
\SetKwData{TurnMagnitudes}{turn\_magnitudes}
\SetKwData{Peaks}{peaks}
\SetKwData{Prominences}{prominences}
\Input{
a list of images $I_1 \dots I_n \in \I$ \\
max number of turns \NumTurns

} 

\Output{a list of \TurnCenters}
\BlankLine
\TurnCenters $\leftarrow$ []\;
\PotentialTurns $\leftarrow$ []\;
\TurnMagnitudes $\leftarrow$ []\;

\BlankLine
\PotentialTurns, \TurnMagnitudes $\leftarrow$ \ComputePotentialTurns{\I}\;

\TurnMagnitudes $\leftarrow$ \MedianFilter(\TurnMagnitudes)\;
\Peaks, \Prominences $\leftarrow$ \FindPeaks{\TurnMagnitudes}\;
\Peaks.\Sort{\Prominences}\;
\Peaks.\Reverse{}\;
\Peaks $\leftarrow$ \Peaks[1 : \Min{\NumTurns, \Len{\Peaks}}]\;
\TurnCenters $\leftarrow$ \PotentialTurns[\Peaks]\;

\KwRet{\TurnCenters}
\caption{Turn Detection}
\label{alg:turn_detection}
\end{algorithm}

\begin{figure*}[htbp]
\centering
  \includegraphics[width=\linewidth]{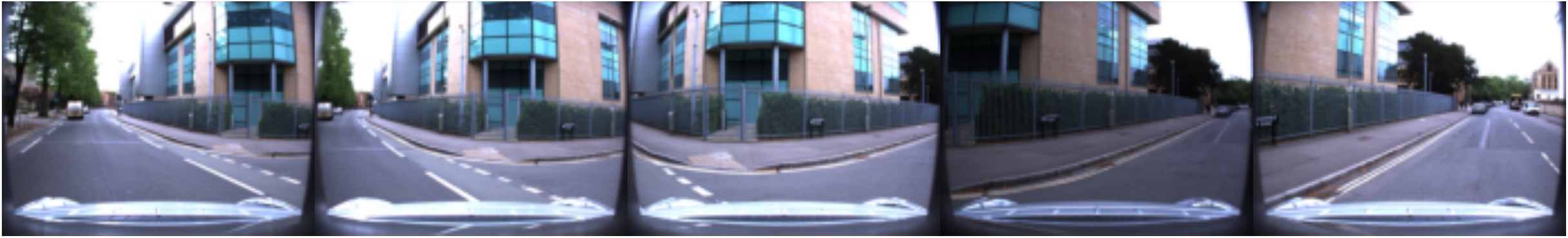}
   \caption{Sample images from a turn detected in the Oxford Robotcar dataset using Algorithm \ref{alg:turn_detection}. The middle image corresponds to the detected turn median. For visualization, the remaining shown images are selected with a step size of 5 frames from the sub-sequence.}
\label{fig:turn}
\end{figure*}

\begin{figure}[tbh]
\centering
  \includegraphics[width=\linewidth]{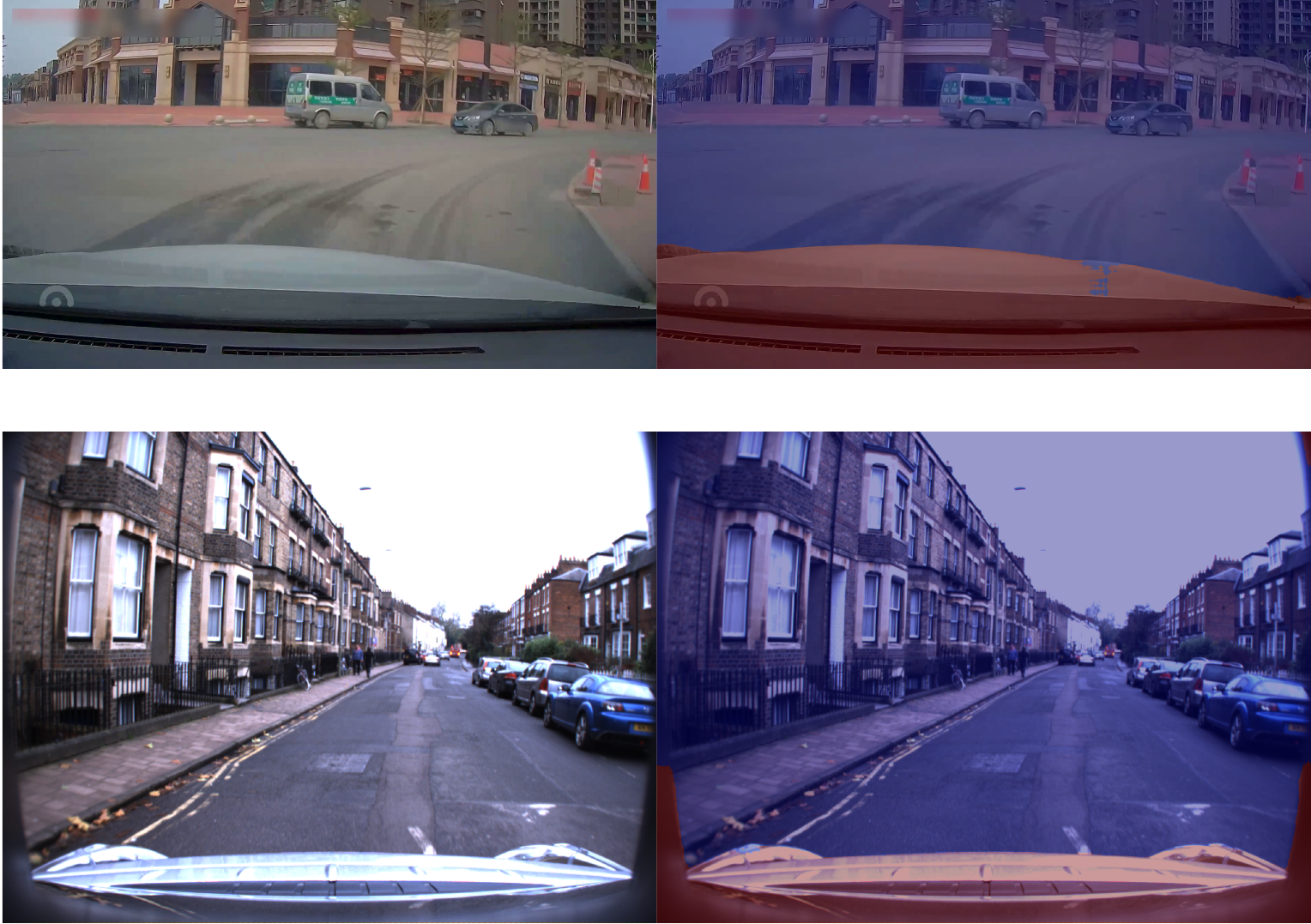}
   \caption{Masked ego-vehitcle using \cite{arani2019rgpnet}. Left: Input image. Right: Segmented image with masked ego-vehicle (red) and valid region (blue). Examples from D$^2$-City (top), and Oxford Robotcar (bottom) datasets. Note the varying shape, angles, and amount of ego-vehicle visibility. Only the valid region is used during feature extraction and matching for calibration.}
\label{fig:masked-ego-vehicle}
\end{figure}

To identify potential turns in the complete image sequence, we utilize a heuristic based on epipoles~\cite{hartley2003multiple}. 
The image sequence is broken into triplets of consecutive images, and  the epipoles are computed for the $C^3_2$ combinations of image pairs. The portions where the vehicle is undergoing pure translation is indicated by the failure to extract the epipoles. For the remaining portions, the average relative distance of the epipoles from the image centers is used as an indicator for the turning magnitude.
We filter the estimated potential turn magnitudes across the sequence for noise, using a median filter with window size $2\cdot \floor{k/2}-1$.

Thereafter, we compute the peaks (local maximas) and the topographic prominences of the potential turn magnitudes. The prominences, sorted in a descending order are the proxy for the likelihood of each peak being a usable turn. Selecting the top $\varsigma$ turns based on the peak prominences, we construct the sub-sequences containing $2k + 1$ images each. Figure \ref{fig:turn} shows a set of sample images from a detected turn in the Oxford Robotcar dataset.

\subsection{Calibration}
\label{sec:calibration}
Given the $\varsigma$ turn sub-sequences, the camera is calibrated within a SfM framework. Each turn is incrementally reconstructed, and the models for the turns are consequently merged to output the final camera parameters as shown in Figure \ref{fig:CalibFlowchart}. Utilizing multiple turns assists in increasing the reliability of the calibration and accounting for any false positive turn-detections that may not be successfully reconstructed. 

Calibration through SfM involves building a scene graph for each turn. The images in the turn are the nodes of this graph, and the validated image pairs with their inlier feature correspondences are the edges. This requires extracting features from the images and matching them across the sequence, followed by a geometric verification of the matches.

\paragraph{Ego-Vehicle Masking:} One of the challenges in crowdsourced dashcamera sequences is the presence of the ego-vehicle in the images. This impacts the correspondence search negatively~\cite{schonberger2016structure}. Upon sampling several crowdsourced sequences, we find that they have varying segments of car dashboard as well as A-pillars visible in the camera view, making a fixed crop of the images an infeasible solution. Therefore, we utilize a real-time semantic segmentation network \cite{arani2019rgpnet} trained on Mapillary dataset \cite{neuhold2017mapillary} for masking out the ego-vehicle (see Figure \ref{fig:masked-ego-vehicle}) during feature extraction.

\paragraph{Reconstruction:} After building the scene graph, we reconstruct each turn within an incremental SfM framework to output a set of camera poses $\mathcal{P} \in SE(3)$ and the scene points $ \mathcal{X} \in \mathbb{R}^3$~\cite{schonberger2016structure}. 

The Perspective-$n$-Point (P$n$P) problem is solved during turn reconstruction to simultaneously estimate the camera poses, scene structure, and the camera parameters, optimized within a bundle adjustment framework.  This uses the correspondences between the 2D image features and the observed 3D world points for image registration. 

Based on this, we show that the error in estimating the focal lengths is bounded by the inverse of the rotation between the matched frames.
For simplicity of derivation, we 
keep the principal point fixed at the image center. 
The camera model maps the observed world points ${P_w}$ to the feature coordinates ${p_c}$ of the rectified image such that,
\begin{equation}
\label{eq:perspective_model}
    sp_c = K R_{cw}P_w + K t_{cw},
\end{equation}
where the product of the camera intrinsic matrix $K$ (Eq. \ref{eq:camera_matrix}) and the rigid camera pose with rotation $R_{cw}\in SO(3)$ and translation $t_{cw} \in \mathbb{R}^3$ is the projection matrix, and $s$ denotes scale factor proportional to the projective depth of the features. The distortion model (Eq. \ref{eq:radial_distortion}) further relates the distorted feature coordinates with the rectified feature coordinates $p_c$.
Accordingly, we can associate the feature matches of the observed 3D points in any two ($i,j$) of the $m$ camera views correspondences through 
\begin{equation}
    \label{eq:perspective_projected_model}
    sp_j = KRK^{-1}p_i + Kt.
\end{equation}

In the scenario of pure translation without any rotation, Eq. \ref{eq:perspective_projected_model} reduces to 
\begin{equation}
    \label{eq:pure_translation}
    sp_j = Kt,
\end{equation}
showing that the correspondences cannot be utilized for calibration. 
Now consider the scenario where there is no translation. The feature/pixel shift is then only determined by the amount of rotation,
\begin{equation}
\label{eq:perspective_only_rotation}
    p_j = \dfrac{KRK^{-1}p_i}{\left(KRK^{-1}p_i\right)_3}, 
\end{equation}
where the subscript $3$ represents the third component of the $3\times1$ homogeneous feature coordinates vector. Having correspondences across overlapping frames allows for the assumption that relative rotation is small. Hence, we can write
\begin{align}
\label{eq:small_rotation}
    R &= \mathcal{I} + r, \\
    \text{where} \
    r &= \begin{bmatrix}
    0 && r_z && -r_y \\
    -r_z && 0 && r_x \\
    r_y && -r_x && 0
    \end{bmatrix}.
\end{align}
To be able to finally derive an analytic expression, we expand Eq. \ref{eq:perspective_only_rotation} for $r$  using Taylor series to obtain,
\begin{equation}
    \label{eq:taylor_series}
    p_j = p_i + (KrK^{-1}p_i) - p_i(KrK^{-1}p_i)_3,
\end{equation}
where the subscript $3$ represents the  third component of the $3\times1$ homogeneous feature coordinates vector. Since vehicular motion primarily consists of the car turning about the $y$ axis, we only consider the yaw $r_y$ for this derivation. 
Substituting the camera matrix from Eq. \ref{eq:camera_matrix}
and the rotation matrix from Eq. \ref{eq:small_rotation} in Eq. \ref{eq:taylor_series}, we obtain
\begin{equation}
\label{eq:pcx_expanded}
    p_{j(x)} = -f_x r_y - r_y  \dfrac{(p_{i(x)} - c_x)^2}{f_x},
\end{equation}
\begin{equation}
\label{eq:pcy_expanded}
    p_{j(y)} = -r_y\dfrac{(p_{i(x)} - c_x)(p_{i(y)} - c_y)}{f_x}.
\end{equation}
Similar equations can also be written for the estimated camera parameters and the reprojected feature coordinates. 
Bundle adjustment minimizes the re-projection error,
\begin{align}
    \label{eq:bundle_adjust}
    \varepsilon &= \sum_{j}\rho_j \left(\norm{\hat{p}_j - p_j}_2^2 \right) \\
    &= \sum_{j}\rho_j \left(\lvert\delta p_{j(x)}\rvert^2 + \lvert\delta p_{j(y)}\rvert^2\right),
\end{align}
where $\rho_j$ down-weights the outlier matches, and $\hat{p}_j = p_j + \delta p_j$, are the reprojected feature coordinates. 
Thus, the parameters to be estimated, focal lengths $\hat{f}_x, \hat{f}_y$, camera rotation $\hat{r}_y$, and distortion coefficients $k_1$, $k_2$ appear implicitly in the error function above. 

Since the camera rotation and intrinsics are optimized simultaneously to minimize the re-projection error, we can assume that the estimated $\hat{r}_y$ balances the estimated camera parameters. For simplicity, we choose a yaw that at least the features close to the principal point remain unchanged, where the impact of distortion is also negligible. Therefore, from Eqs. \ref{eq:pcx_expanded} and \ref{eq:pcy_expanded}, we understand
\begin{equation}
\label{eq:rotation_focal_relation}
    \hat{r}_y\hat{f}_x = r_yf_x.
\end{equation}
Now we can write the equations for the estimated feature point coordinates replacing $\hat{r}_y$ to obtain,
\begin{equation}
\label{eq:pcx_hat_expanded}
    \hat{p}_{j(x)} = -f_x r_y - f_x r_y \dfrac{(p_{i(x)} - c_x)^2}{\hat{f}^2_x},
\end{equation}
\begin{equation}
\label{eq:pcy_hat_expanded}
    \hat{p}_{j(y)} = -f_x r_y\dfrac{(p_{i(x)} - c_x)(p_{i(y)} - c_y)}{\hat{f}^2_x}.
\end{equation}
Accordingly, by subtracting Eqs. \ref{eq:pcx_expanded} and \ref{eq:pcy_expanded} from Eqs. \ref{eq:pcx_hat_expanded} and \ref{eq:pcy_hat_expanded} respectively, we get the first order approximation of $\delta p_j$,
\begin{equation}
\label{eq:delta_px}
   \delta p_{j(x)} = 2\delta f_x r_y \dfrac{(p_{i(x)} - c_x)^2}{f^2_x},
\end{equation}
\begin{equation}
\label{eq:delta_py}
   \delta p_{j(y)} = 2\delta f_x r_y \dfrac{(p_{i(x)} - c_x)(p_{i(y)} - c_y)}{f^2_x}.
\end{equation}
For the errors $\lvert \delta p_j(x) \rvert \ll 1$ and $\lvert \delta p_j(y) \rvert \ll 1$, we need to simultaneously satisfy
\begin{align}
\label{eq:err_fx}
    \lvert \delta f_x \rvert &\ll \dfrac{f^2_x}{2r_y (p_{i(x)} - c_x)^2} \ \text{ and},\\
    \lvert \delta f_x \rvert &\ll \dfrac{f^2_x}{2r_y \lvert p_{i(x)} - c_x \rvert \lvert p_{i(y)} - c_y \rvert }
\end{align}
with the features closer to the boundary providing the tightest bound on $\lvert \delta f_x \rvert$. Since $c_x = w/2$ and $c_y = h/2$, we obtain,

\begin{equation}
\label{eq:calibration_error_bound}
\begingroup 
\setlength\arraycolsep{2pt}
\begin{bmatrix}
    \delta{f_x} &&
    \delta{f_y}
\end{bmatrix}^\intercal
< 
\dfrac{2}{\text{max}(h,w)}
\begin{bmatrix}
\dfrac{f_x^2}{w\cdot r_y} &&
\dfrac{f_y^2}{h\cdot r_x}
\end{bmatrix}^\intercal.
\endgroup
\end{equation}
This result is similar to the Eq. 3 obtained in \cite{gordon2019depth}.
Thus the errors in focal lengths $f_x$ and $f_y$ are bounded by the inverse of rotation about the $y$ and $x$ axes, respectively. 

Since the vehicular motion is planar with limited pitch rotations, we perform the calibration in two steps:
\begin{enumerate}
    \item As modern cameras are often equipped with nearly square pixels, we fix $f_x = f_y$ and utilize the yaw rotations about the $y$ axis to calibrate the focal lengths. The principal point is also fixed to the center of the image. This step outputs a common focal length and the refined distortion coefficients.
    \item Thereafter, we relax these assumptions and refine all the parameters simultaneously within bounds. This allows to slightly update $f_y$ and also refine the principal point to minimize the reprojection error.
\end{enumerate}
The final output is a set of six parameters namely, focal lengths $f_x$, and $f_y$, principal point ($c_x$, $c_y)$, and the radial distortion coefficients $k_1$ and $k_2$.
Thereafter, the reconstructed models of the turns are merged. Thus, instead of optimizing the intrinsics as separate parameters for each turn, they are estimated as a single set of constants. In the few scenarios where overlapping turn sub-sequences may be present, they are also spatially merged and optimized together. 

\begin{table*}[tbhp]
\centering
\resizebox{0.85\textwidth}{!}{%
\begin{tabular}{|c|c|cccccc|c|}
\hline
\textbf{Dataset} & \textbf{Method} &\cellcolor{red!25} $\downarrow\mathbf{f_x}$ &\cellcolor{red!25} $\downarrow\mathbf{f_y}$ &\cellcolor{red!25} $\downarrow\mathbf{c_x}$ &\cellcolor{red!25} $\downarrow\mathbf{c_y}$ &\cellcolor{red!25} $\downarrow\mathbf{k_1}$ &\cellcolor{red!25} $\downarrow\mathbf{k_2}$ &\cellcolor{blue!25} $\uparrow\mathbf{SSIM}_\mu$ \\ \hline \hline

\multirow{2}{*}{\textbf{\begin{tabular}[c]{@{}c@{}}KITTI\\ Raw \end{tabular}}} & \cite{chawla2020monocular} & 1.735 & 0.967 & 0.884 & \colorbox{lightergray}{0.051} & 11.435 & 34.488 & - \\
 & Ours (w/o M) & \colorbox{lightergray}{1.670} & \colorbox{lightergray}{0.525} & \colorbox{lightergray}{0.717} & 0.522 & \colorbox{lightergray}{11.421} & \colorbox{lightergray}{34.034} & - \\ \hline

\multirow{3}{*}{\textbf{\begin{tabular}[c]{@{}c@{}}Oxford\\ (Good GPS)\end{tabular}}}  
 & \cite{chawla2020monocular} & \colorbox{lightergray}{7.065} & \colorbox{lightergray}{6.018} & 0.715 & \colorbox{lightergray}{0.151} & - & - & 0.889 \\
& Ours (w/o M) & 8.068 & 6.394 & 1.152 & 0.433 & - & - & 0.890 \\ 
& Ours (with M) & 7.598 & 6.286 & \colorbox{lightergray}{0.587} & 0.426 & - & - & \colorbox{lightergray}{0.893} \\ \hline

\multirow{3}{*}{\textbf{\begin{tabular}[c]{@{}c@{}}Oxford \\ (Poor GPS)\end{tabular}}} 
 & \cite{chawla2020monocular} & 6.763 & 6.156 & 0.418 & \colorbox{lightergray}{0.017} & - & - & 0.908 \\
 & Ours (w/o M) & 1.975 & 0.703 & 0.513 & 3.010 & - & - & 0.906 \\ 
 & Ours (with M) & \colorbox{lightergray}{1.892} & \colorbox{lightergray}{0.361} & \colorbox{lightergray}{0.386} & 2.516 & - & - & \colorbox{lightergray}{0.909} \\\hline
 
\multirow{2}{*}{\textbf{\begin{tabular}[c]{@{}c@{}}Oxford \\ (No GPS)\end{tabular}}} & Ours (w/o M) & 13.937 & 18.956 & \colorbox{lightergray}{0.857} & 1.639 & - & - & 0.897 \\ 
& Ours (with M) & \colorbox{lightergray}{6.610} & \colorbox{lightergray}{6.408} & 1.207 & \colorbox{lightergray}{0.549} & - & - & \colorbox{lightergray}{0.899} \\\hline
\end{tabular}}
\caption{Comparing calibration performance measured through median absolute percentage error and mean SSIM across datasets. M represents ego-vehicle masking. The best estimates for each parameter on the datasets are highlighted in gray.}
\label{tab:calibration_performance_comparison}
\end{table*}

\section{\uppercase{Experiments}}
\label{sec:experiments}
\noindent We validate our proposed system for auto-calibration of onboard cameras on the KITTI raw \cite{geiger2012we}, the Oxford Robotcar \cite{maddern20171}, and the D$^2$-city \cite{che2019d} datasets. Ego-vehicle is visible in the onboard camera of Oxford Robotcar and the dashcamera of D$^2$-City datasets, respectively. Corresponding GPS information (used in competing methods) is not availbale for the crowdsourced  D$^2$-City dataset.

We compare our system against \cite{chawla2020monocular} which uses GPS based turn detection and SfM for calibration, as well as \cite{santana2017automatic} which relies upon detecting lines, curves and Hough transform in the scene. We also demonstrate the necessity of masking the ego-vehicle for accurate calibration.
To further evaluate our proposed system, 
we empirically study the impact of the number of turns used, as well as the frame rate of the videos on the calibration performance. We show that our system is superior to \cite{santana2017automatic} as well as \cite{chawla2020monocular,chawla2020crowdsourced}, which in turn outperforms the self-supervised \cite{gordon2019depth}. Finally, we demonstrate the application of our system for chessboard-free accurate monocular dense depth and ego-motion estimation on uncalibrated videos.

\subsection{Datasets}
\paragraph{KITTI raw.} This dataset contains sequences from the urban environment in Germany with a right-hand drive. The images are captured at 10 fps with the corresponding GPS at 100 Hz. The ground truth (GT) camera focal lengths, principal point, and the radial distortion coefficients and model are available for the different days the data was captured. The GT camera parameters corresponding to Seq 00 to 02 are \{$f_x = 960.115 \text{ px}, f_y = 954.891 \text{ px}, c_x = 694.792 \text{ px}, c_y = 240.355 \text{ px}, k_1 = -0.363, k_2 = 0.151$\}, and for Seq 04 to 10 are \{$f_x = 959.198 \text{px}, f_y = 952.932 \text{ px}, c_x = 694.438 \text{ px}, c_y = 241.679 \text{ px}, k_1 = -0.369, k_2 = 0.158$\}. Seq 03 is not present in the raw dataset. Also, the ego-vehicle is not visible in the captured data.

\noindent\paragraph{Oxford Robotcar.} This dataset contains sequences from  the urban environment in the United Kingdom with a left-hand drive. The images are captured at 16 fps, and the GPS at 16 Hz. However, some of the sequences have poor GPS or even do not have corresponding GPS available. A single set of GT camera focal lengths and the principal point is available for all the recordings, \{$f_x = 964.829 \text{ px}, f_y = 964.829 \text{ px}, c_x = 643.788 \text{ px}, c_y = 484.408 \text{ px}$\}.
Instead of the camera distortion model and coefficients, a look-up table (LUT) is available that contains the mapping between the rectified and distorted images. The ego-vehicle is also visible in the captured data.

\noindent\paragraph{D$^2$-City.} This dataset contains crowdsourced sequences collected from dashboard cameras onboard DiDi taxis in China. Therefore, the ego-vehicle is visible in the captured data. Different sequences have different amount and shape of this dashboard and A-pillar visibility. There is no accompanying GPS. The images are collected at 25 fps across varying road and traffic conditions. Consequently, no GT camera parameters are available as well.

\begin{figure*}[tbh]
\centering
  \includegraphics[trim=0 0 0 0cm, width=\linewidth]{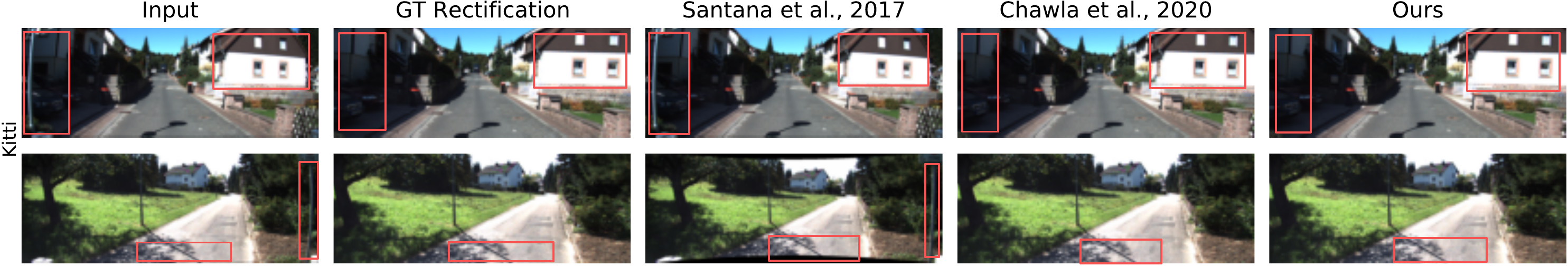}
  \includegraphics[trim=0 0 0 0cm, width=\linewidth]{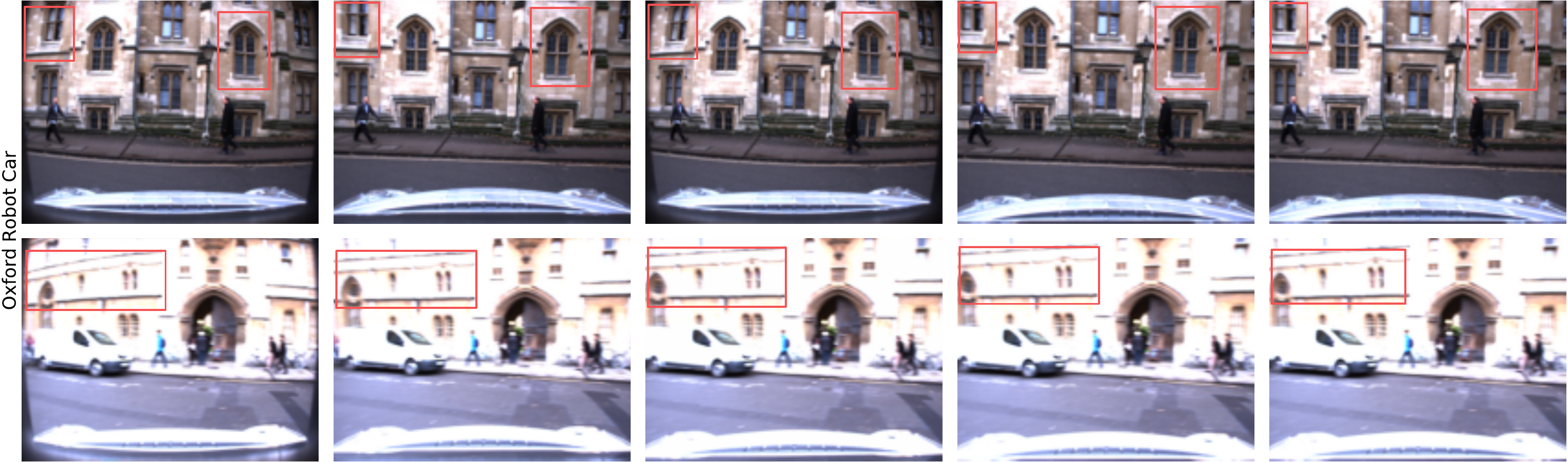}
  \includegraphics[trim=0 0 0 0cm, width=\linewidth]{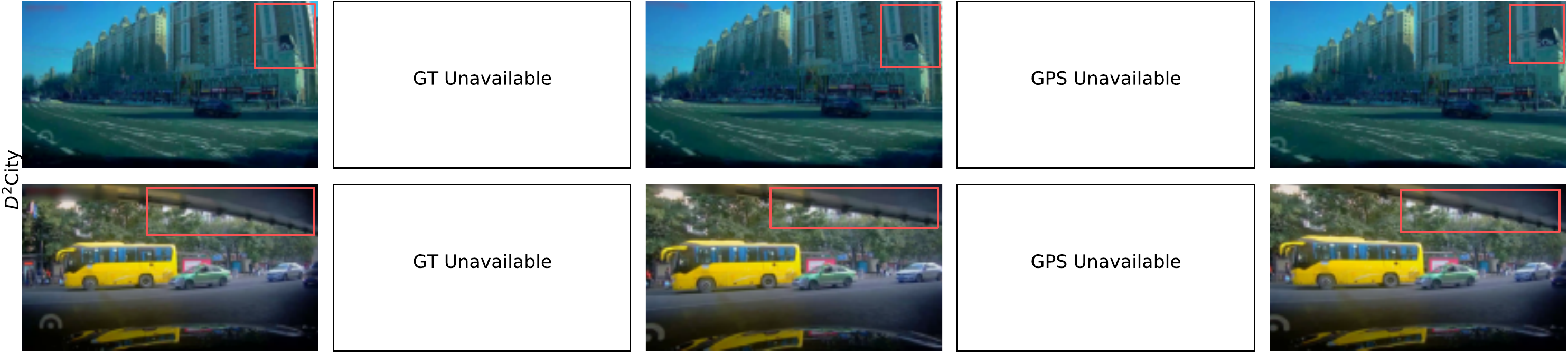}
  
   \caption{Qualitative comparison of camera auto-calibration on the KITTI raw (top), the Oxford Robotcar (middle), and the D$^2$-city (bottom) datasets. 
   }
\label{fig:qual_performance}
\end{figure*}

\begin{figure}[tbh]
\centering
  \includegraphics[width=\columnwidth]{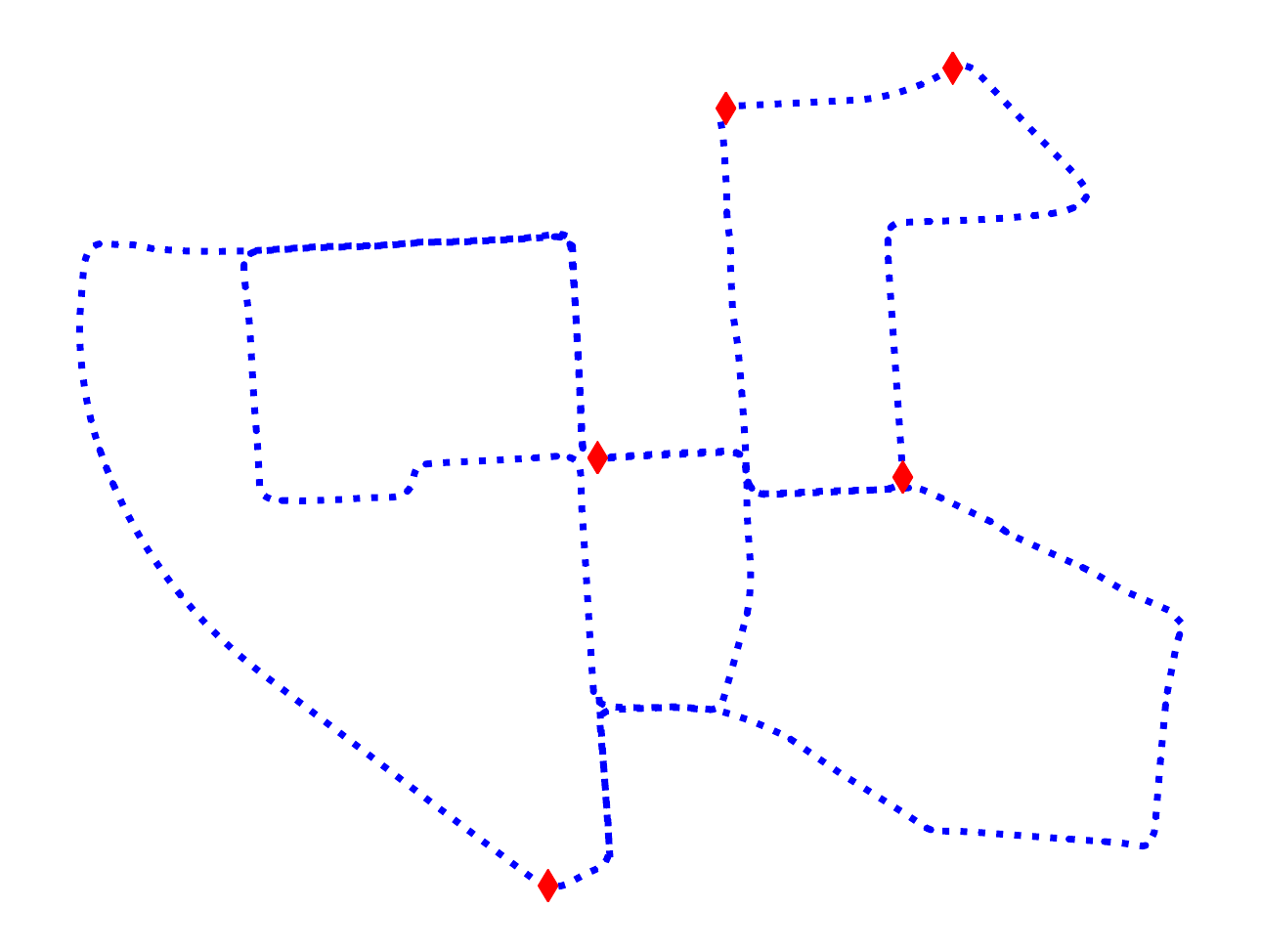}
  \caption{Using the top 5 turns for auto-calibration of KITTI Seq 00. The red diamonds denote the turn centers, and the dotted blue line is the corresponding GPS trajectory. Note that our method does not use this GPS trajectory for turn estimation. 
  }
\label{fig:top_turns}
\end{figure}

\begin{figure*}
  \centering
  \includegraphics[width=0.41\textwidth]{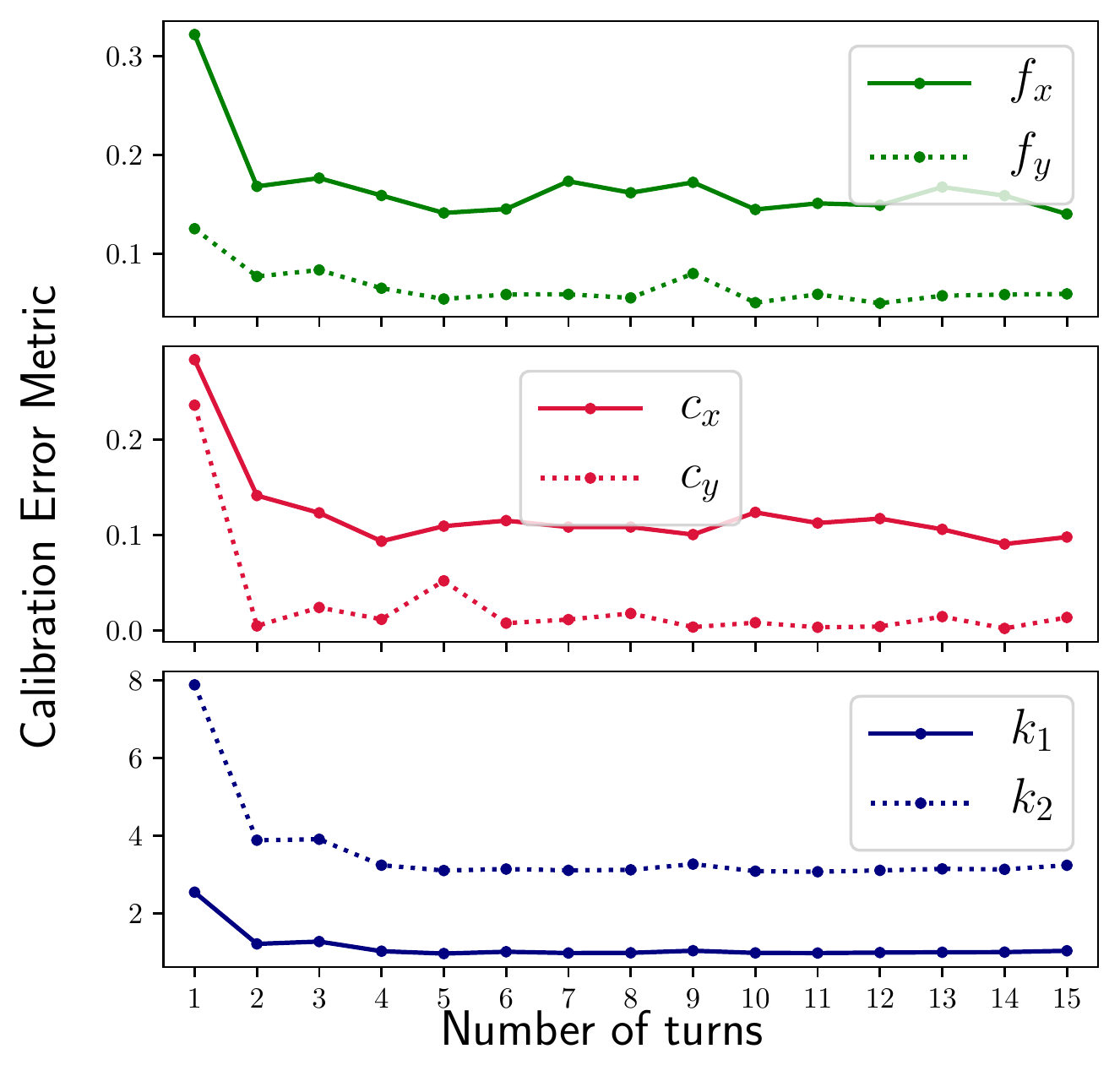}
  \includegraphics[width=0.41\textwidth]{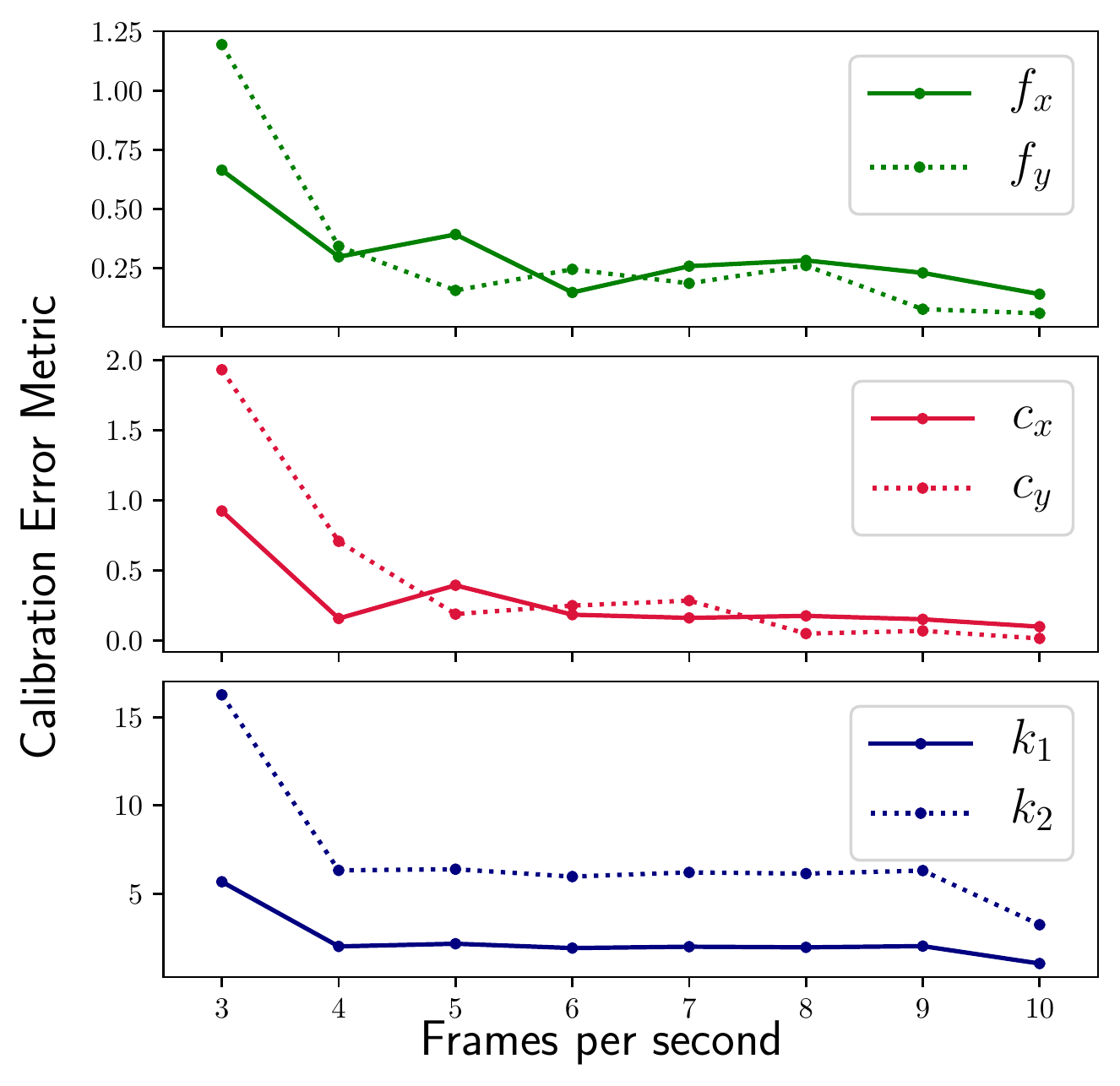}
  \caption{Impact of the number of turns used and the frame rate of captured image sequence on the calibration performance.  The calibration error metric (unitless) measures the median of absolute percentage error normalized by the number of times the calibration was successful.
  }
\label{fig:turns_plot}
\end{figure*}

\subsection{Performance Evaluation}
Table \ref{tab:calibration_performance_comparison} summarizes the results of performance evaluation of our proposed onboard monocular camera auto-calibration system.
We evaluate on ten KITTI raw sequences 00 to 10 (except sequence 03, which is missing in the dataset), and report the average calibration performance. Furthermore, we evaluate on three sequences from the Oxford Robotcar dataset with different GPS qualities, \texttt{2014-11-28-12-07-13} with good GPS measurements, \texttt{2015-03-13-14-17-00} with poor GPS measurements, and \texttt{2015-05-08-10-33-09} without any accompanying GPS measurements. We repeat each calibration 5 times are report the median absolute percentage error for the estimated focal lengths, principal point, and the distortion coefficients. Since the GT distortion model and coefficients are not provided in the Oxford robotcar dataset, we instead measure the mean structural similarity \cite{wang2004image} between the images rectified using the provided Look-up-table (LUT) and our estimated parameters. Finally, we calibrate 10 turn-containing sequences of \SI{30}{s} each, from the  D$^2$-City dataset. 

We successfully calibrate all sequences except for KITTI 04 which does not contain any turns. Our method performs better than \cite{chawla2020monocular} in all the cases except for Oxford (Good GPS).  This case can be attributed to better turn detection with good GPS quality used by that method, which is often not available for crowdsourced data. 
Also, note that the calibration is better when using the proposed ego-vehicle masking. This effect is most pronounced for Oxford (No GPS) where the focal length errors become nearly half when masking out the ego-vehicle. The calibration without ego-vehicle masking is less effective with the presence of ego-vehicle in the images, as it acts as a watermark and negatively impacts feature matching \cite{schonberger2016structure}, even with the use of  RAndom SAmple Consensus (RANSAC)~\cite{hartley2003multiple}. 

Following the comparison protocol of \cite{santana2017automatic}\footnote{\url{http://dev.ipol.im/~asalgado/ipol_demo/workshop_perspective/}}, Figure \ref{fig:qual_performance} shows some qualitative auto-calibration results comparing our system against \cite{chawla2020monocular,santana2017automatic}.  Note that our system performs better, as visibly demonstrated by the rectified structures. Moreover, our method is applicable even when no GPS information is available, as we successfully calibrate all the selected sequences from D$^2$-City. However, our system relies upon the features in the images and thus is more suitable for urban driving sequences. Additional qualitative results for the dashcamera videos from D$^2$-city can be found in the Appendix.

\subsection{Design Evaluation}
Here, we further evaluate our proposed system design for the impact of the number of turns used during calibration, and the frame rate of the onboard image sequence, on the calibration performance. 
We carry out these analyses on KITTI sequence 00. For all these experiments the value of $k$ is set to 30. 

For each experiment repeated multiple times, we report the \textit{calibration error metric} as the median of absolute percentage error normalized by the number of times the calibration was successful. This unitless metric is used to capture the accuracy as well as the consistency of the auto-calibration across the aforementioned settings.

\begin{figure*}[tbh]
\centering
 \includegraphics[width=\linewidth]{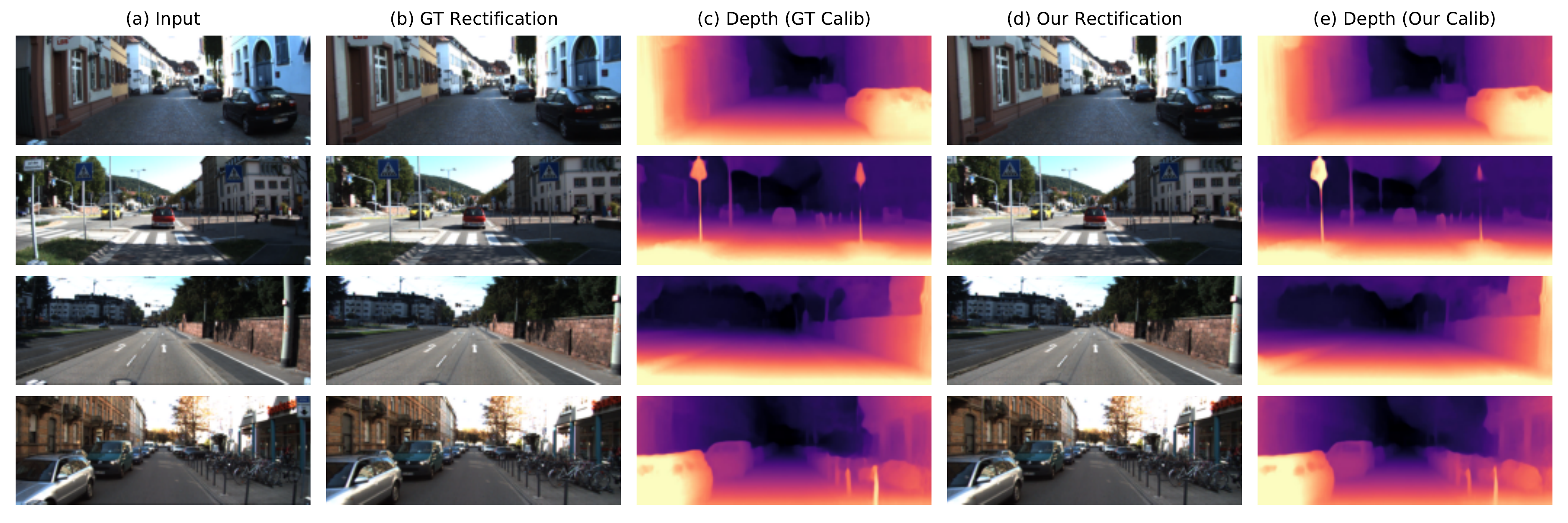}
   \caption{Comparison of monocular dense depth estimation when training Monodepth-2 using GT and our estimated camera parameters.}
\label{fig:depth_maps}
\end{figure*}

 \paragraph{Number of Turns:}
\label{sec:number_of_turns}
Here, we study the impact of the number of turns on the calibration error metric. For this analysis, we first create a set $S_{turns}$ of the top 15 turns (at 10 fps) extracted using Algorithm \ref{alg:turn_detection}.
Thereafter, we vary the number of turns $j$ used for calibration by randomly selecting them from $S_{turns}$. 
We repeat each experiment 10 times and report the calibration error metric (see Figure \ref{fig:turns_plot}).
Note that the focal lengths estimation improves up to two turns. The principal point estimation improves up to four turns. The distortion coefficients estimation improves up to five turns. 
Therefore, successful auto-calibration with our approach requires a minimum of 5 turns, consisting of only a few hundred ($\approx$300) images. Figure \ref{fig:top_turns} shows the top 5 of the extracted turns for auto-calibration of KITTI Sequence 00. This reinforces the practicality and scalability of our system.

 \paragraph{Frame Rate:}
We study the impact of the frame rate by sampling the image sequence across a range of 1 fps to 10 fps, and report the calibration metric as before (see Figure \ref{fig:turns_plot}). Note that calibration is unsuccessful when the frame rate is less than 3 fps.
Thereafter, the calibration parameters improve up to 5 fps, thereby demonstrating the efficacy of our method even for low-cost onboard cameras.

\begin{table}[b]
\centering
\resizebox{\columnwidth}{!}{
\begin{tabular}{|l|cc|c|}
\hline
\textbf{Calib} & \cellcolor{red!25}$\downarrow$\textbf{Abs Rel Diff} & \cellcolor{red!25}$\downarrow$\textbf{RMSE} & \cellcolor{blue!25}$\uparrow$\bm{$\delta < 1.25$} \\ \hline \hline
\textbf{\cite{chawla2020monocular}} & 0.1142 & 4.8224 & 0.8783 \\ 
\textbf{Ours} & \colorbox{lightergray}{0.1137} &  \colorbox{lightergray}{4.7895} & \colorbox{lightergray}{0.8795} \\ 
\hline
\end{tabular}}
\caption{Chessboard-free monocular dense depth estimation with metrics defined in \cite{zhou2017unsupervised}. 
Better results are highlighted.}
\label{tab:depth}
\end{table}

\subsection{Spatial Scene-Understanding on Uncalibrated Sequences}
We apply our auto-calibration system for training monocular dense depth and ego-motion estimation on the KITTI Eigen \cite{eigen2015predicting} and Odometry splits \cite{zhou2017unsupervised} respectively, without prior knowledge of the camera parameters. Tables \ref{tab:depth} and \ref{tab:ego-motion} compare depth and ego-motion estimation using Monodepth-2 \cite{godard2018digging} with different calibrations on the metrics defined in \cite{zhou2017unsupervised}, respectively. Note that depth and ego-motion estimation using our parameters is better than that using \cite{chawla2020monocular}. 
Figure \ref{fig:depth_maps} provides some qualitative examples of depth maps estimated using our proposed system, which are comparable to those when using GT calibration.

\begin{table}[t!]
\centering
\resizebox{\columnwidth}{!}{
\begin{tabular}{|c|c|c|}
\hline
\textbf{Calib} & \textbf{Seq 09} & \textbf{Seq 10} \\ \hline \hline
\textbf{\cite{chawla2020monocular}} & 0.0323 $\pm$ 0.0103 & 0.0228 $\pm$ 0.0132 \\
\textbf{Ours} & \colorbox{lightergray}{0.0299 $\pm$ 0.0109} & \colorbox{lightergray}{0.0210 $\pm$ 0.0125} \\ \hline
\end{tabular}}
\caption{Absolute Trajectory Error (ATE-5) \cite{zhou2017unsupervised} for chessboard-free monocular ego-motion estimation on KITTI sequences 09 and 10. 
Better results are highlighted.}
\label{tab:ego-motion}
\end{table}

\section{\uppercase{Conclusions}}
\noindent In this work, we demonstrated spatial-scene understanding through practical monocular camera auto-calibration of crowdsourced onboard or dashcamera videos. Our system utilized the structure reconstruction of turns present in the image sequences for successfully calibrating the KITTI raw, Oxford robotcar, and the D$^2$-city datasets.
We showed that our method can accurately extract these turn sub-sequences of total length $\approx$ \SI{30}{s} from long videos  themselves, without any assistance of corresponding GPS information. 
Moreover, our method is effective even on low fps videos for low-cost camera applications. Furthermore, the calibration performance was improved by automatically masking out the ego-vehicle in the images.
Finally, we demonstrated chessboard-free monocular dense depth and ego-motion estimation for uncalibrated videos through our system. Thus, we contend that our system is suitable for utilizing crowdsourced data collected from low-cost setups to accelerate progress in autonomous vehicle perception at scale.

\bibliographystyle{apalike}
{\small
\bibliography{ref.bib}}

\section*{\uppercase{Appendix}}
\label{sec:appendix}
\begin{figure*}[htbp]
\centering
  \includegraphics[width=0.9\linewidth]{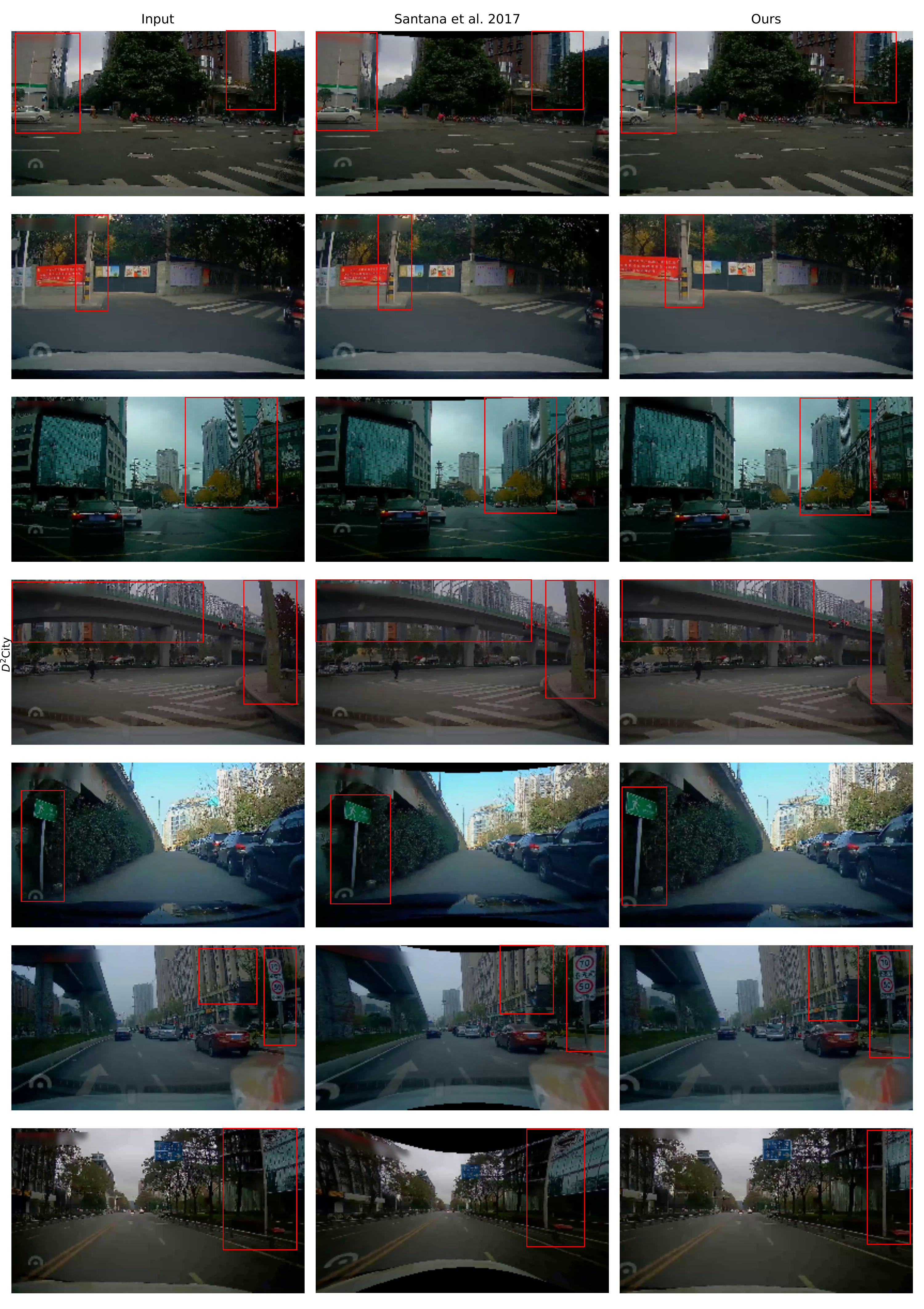}
  
   \caption{Additional qualitative comparison of camera auto-calibration on the D$^2$-city dataset. 
   }
\label{fig:qual_performance_append}
\end{figure*}

We provide additional qualitative results on the calibration of the $D^2$-City dataset.
Here, we don't show the results from \cite{chawla2020monocular} because it requires corresponding GPS headings that are missing for this dataset. 
Note that \cite{santana2017automatic} is run with their default parameters. As shown in Fig. \ref{fig:qual_performance_append}, sometimes \cite{santana2017automatic} fails to completely undistort the structures, while our method performs consistently across a variety of real-world situations. Note that the average estimated focal length for D$^2$-City cameras is $\approx 1400 \text{ px}$, different from the range of values for the KITTI and the Oxford Robotcar datasets.

\end{document}